\pdfoutput=1

\documentclass[11pt]{article}

\usepackage[final]{acl}

\usepackage{times}
\usepackage{latexsym}

\usepackage[T1]{fontenc}

\usepackage[utf8]{inputenc}

\usepackage{microtype}

\usepackage{inconsolata}

\usepackage{graphicx}

\usepackage{algorithmic}
\usepackage{algorithm}
\usepackage{amssymb}
\usepackage{amsmath}
\usepackage{xcolor}
\usepackage{tikz}
\usetikzlibrary{calc}
\usetikzlibrary{matrix}
\usetikzlibrary{shapes.arrows}
\usetikzlibrary{decorations.pathreplacing,calligraphy}
\usetikzlibrary{patterns}
\usepackage{pgfplots}
\usepackage{pgfplotstable}
\usepackage{subcaption}
\usepackage{multirow}
\usepackage{enumitem}
\definecolor{darkteal}{HTML}{2E5F7F}

\let\oldequation\equation
\let\oldendequation\endequation
\renewenvironment{equation}{\small\oldequation}{\oldendequation}

\newcommand{\paperName}{KVPR}

\title{\paperName: Efficient LLM Inference \\ with I/O-Aware \underline{KV} Cache \underline{P}artial \underline{R}ecomputation}

\author{%
  Chaoyi Jiang$^{*}$, Lei Gao\thanks{These authors contributed equally.}, Hossein Entezari Zarch, Murali Annavarm\\
  University of Southern California\\
  \texttt{\{chaoyij, leig, entezari, annavara\}@usc.edu}\\
}

\begin{document}
\maketitle

\begin{abstract}
Inference for Large Language Models (LLMs) is computationally demanding. To reduce the cost of auto-regressive decoding, Key-Value (KV) cache is used to store intermediate activations, which significantly lowers the computational overhead for token generation. However, the memory required for the KV cache grows rapidly, often exceeding the capacity of GPU memory. A cost-effective alternative is to offload KV cache to CPU memory, which alleviates GPU memory pressure, but shifts the bottleneck to the limited bandwidth of the PCIe connection between the CPU and GPU. Existing methods attempt to address these issues by overlapping GPU computation with I/O or employing CPU-GPU heterogeneous execution, but they are hindered by excessive data movement and dependence on CPU capabilities. Fully overlapping PCIe communication latency gets challenging as the size of the KV cache grows and/or the GPU compute capabilities increase. In this paper, we introduce KVPR, an efficient I/O-aware LLM inference method where the CPU first transfers a partial set of activations, from which the GPU can start recomputing the KV cache values. While the GPU recomputes the partial KV cache, the remaining portion of the KV cache is transferred concurrently from the CPU. This approach overlaps GPU recomputation with KV cache transfer to minimize idle GPU time and maximize inference performance. KVPR is fully automated by integrating a profiler module that utilizes input characteristics and system hardware information, a scheduler module to optimize the distribution of computation and communication workloads, and a runtime module to efficiently execute the derived execution plan. Experimental results show that KVPR achieves up to 35.8\% lower latency and 46.2\% higher throughput during decoding compared to state-of-the-art approaches. The code is available at \url{https://github.com/chaoyij/KVPR}.
\end{abstract}

\section{Introduction}
Large language models (LLMs) have made remarkable progress in recent years, demonstrating their ability to power diverse applications such as machine translation \cite{zhu2024multilingual}, summarization \cite{openai2024gpt4}, creative content generation \cite{geminiteam2024gemini}, and personalized recommendations \cite{geng2022p5}. Real-time applications, including conversational agents and live translation \cite{li2023camel}, depend on low latency to provide seamless user interaction, while large-scale deployments require high throughput to support concurrent users and process substantial data efficiently \cite{kwon2023vllm}.

Key-Value (KV) cache is essential in auto-regressive decoding for LLMs, as it stores the intermediate key and value activations from earlier steps in the attention mechanism. This reduces the computational complexity of generating each token from quadratic to linear by eliminating the need to recompute these activations for every generated token. However, this comes at a cost: the size of the KV cache grows linearly with batch size, sequence length, and model size, leading to substantial memory demands \cite{wan2024efficient}.

GPU memory, while optimized for high-bandwidth access by computation units, is inherently limited and often insufficient to handle the large and growing size of the KV cache. One cost-effective approach to address this limitation is to offload the KV cache to cheaper and plentiful CPU memory, and could be further offloaded to hard disks and network storage \cite{liu2024cachegen}. While offloading reduces GPU memory pressure, it introduces a new bottleneck: the slow PCIe bus becomes a limiting factor when transferring the KV cache from CPU to GPU for computation. Due to the long PCIe transfer time, the overall decoding latency increases and token generation throughput decreases, hindering the overall inference efficiency of the system \cite{zhao2024hetegen}.

\begin{figure}
    \centering
    \includegraphics[width=1\columnwidth]{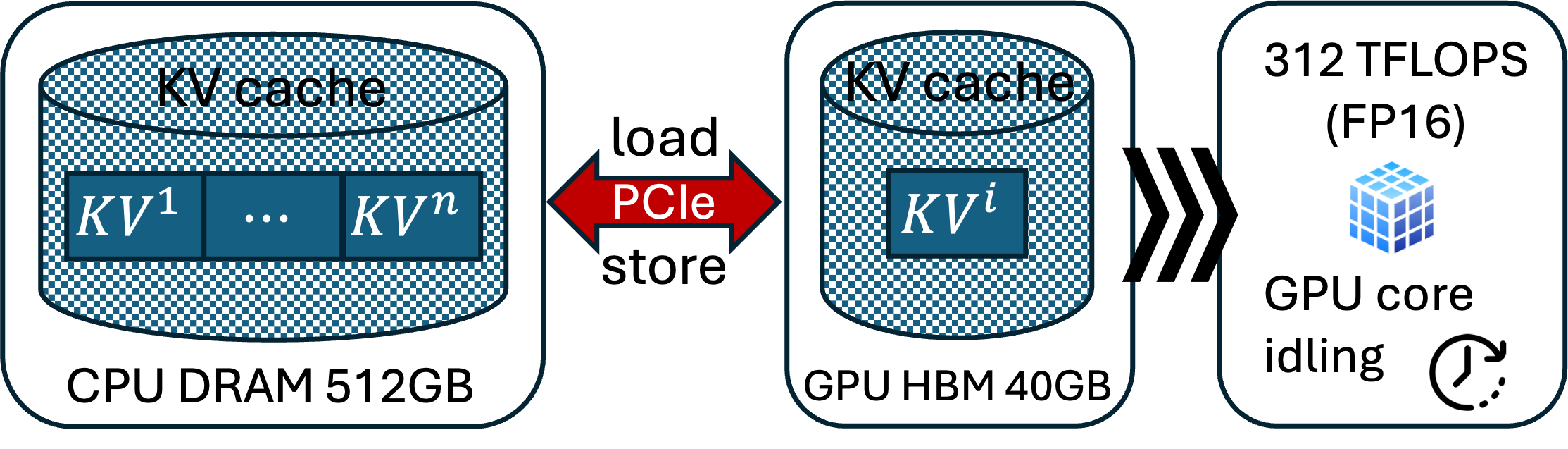}
    \caption{LLM inference system with an A100 GPU.}
    \label{fig:intro}
\end{figure}

\begin{table}[h]
\centering
\resizebox{\columnwidth}{!}{%
\begin{tabular}{lcccc}
\hline
\textbf{Model} & \multicolumn{1}{l}{\textbf{Hidden Dim}} & \textbf{\begin{tabular}[c]{@{}c@{}}KV Cache \\ (MB)\end{tabular}} & \textbf{\begin{tabular}[c]{@{}c@{}}PCIe Latency \\ (ms)\end{tabular}} & \textbf{\begin{tabular}[c]{@{}c@{}}Comp. Latency\\ (ms)\end{tabular}} \\ \hline
OPT-6.7B & 4,096 & 512 & 15.6 & 0.3509 \\
OPT-13B & 5,120 & 640 & 19.5 & 0.4388 \\
OPT-30B & 7,168 & 896 & 27.3 & 0.6143 \\ \hline
\end{tabular}%
}
\caption{PCIe latency and computation latency for different KV cache sizes based on the system in Figure \ref{fig:intro}.}
\label{tab:slow-pcie}
\end{table}

To evaluate the impact of communication overhead, we set up an LLM inference serving system (shown in Figure \ref{fig:intro}) using an NVIDIA A100 GPU. Data transfer between the CPU DRAM and GPU HBM occurs over a PCIe 4.0 16 lanes with a bandwidth of 32 GB/s. Table \ref{tab:slow-pcie} shows the hidden dimension, KV cache size, PCIe transfer time, and GPU computation latency for KV pair computation.  Note that the end-to-end decoding latency includes other components, such as feed-forward layers, which are not shown in this table for clarity. We use FP16 precision with a batch size of 32 and a sequence length of 1024. The results show that PCIe latency exceeds KV cache recomputation latency by over an order of magnitude. Hence, in systems where the KV cache is stored on CPU DRAM the long transfer time leads to GPU idle time, which is detrimental to inference efficiency. 

To mitigate the issue of low latency and bandwidth of PCIe, FlexGen \cite{sheng2023flexgen} and PipeSwitch \cite{bai2020pipeswitch} attempt to overlap GPU computation of the current layer with KV cache loading for the next layer. However, the effectiveness of such an overlap is capped by the task that takes the longest time. In most systems,  PCIe transfer time overshadows GPU computation latency,  particularly with large batch and context sizes. Hence, fully overlapping GPU computation with PCIe transfer time is infeasible. 
FastDecode \cite{he2024fastdecode} suggests computing attention scores directly on the CPU, which has faster access to the KV cache compared to the GPU. Similarly, HeteGen \cite{zhao2024hetegen}, TwinPilots \cite{yu2024twinpilots}, and \citeauthor{park2024improving} employ CPU-GPU heterogeneous execution to hide data transfer overhead by performing computations on the CPU. However, as demonstrated later in our results, such an approach puts a burden on the CPU to satisfy the KV cache computation demands from multiple GPUs attached to a CPU host, thereby limiting scalability.

In this paper, we propose \paperName, a novel approach for efficient LLM inference that balances the GPU computation and PCIe bandwidth tradeoffs. Instead of transferring the entire KV cache from CPU to GPU to compute an attention score, the CPU transfers a partial set of activations, which are smaller in size and are required to generate part of the KV cache, to the GPU. The GPU then starts recomputing the partial KV cache from the input activations. Concurrently, the CPU transfers the remaining KV cache over PCIe. {\paperName} ensures the computation of exact attention scores without approximation, while minimizing GPU idle time and improving overall latency and throughput.

{\paperName} achieves a \textit{near-perfect} overlap of PCIe transfer time and GPU recomputation time by determining the optimal fraction of activations that need to be recomputed. {\paperName} is fully automated in determining the recomputation and communication split. It includes a profiler module that collects system hardware information, a scheduler module that formulates the problem as a linear programming problem to determine the optimal split point, and a runtime module that manages memory allocation on both devices and coordinates data transfer between them. Experimental results show significant improvements in inference latency or throughput, depending on workload. In summary, our contributions are as follows:

\begin{itemize}[topsep=0pt, partopsep=0pt, parsep=0pt, itemsep=2pt]
    \item We propose an efficient CPU-GPU I/O-aware LLM inference method that leverages KV cache partial recomputation with asynchronous KV cache transfer that overlaps compute and communication to address the system bottleneck of loading large KV cache from CPU memories.
    \item We develop a framework based on linear integer programming that achieves optimal computation-communication distribution.
    \item Our experimental results show that {\paperName} outperforms the current state-of-the-art approaches up to 35.8\% in terms of latency and 46.2\% in terms of throughput.
\end{itemize}

\section{Background}

\textbf{LLM inference process.}
The inference process of decoder-only LLMs employs an auto-regressive approach, generating tokens sequentially. It consists of two stages: the \textbf{prefilling stage} and the \textbf{decoding stage}. 
In the prefilling stage, the input to the $ i $-th decoder layer is denoted as $ X^i \in \mathbb{R}^{b \times s \times h} $, where $i \in \{1,\ldots,n\} $, $ b $ is the batch size, $ s $ is the prompt length, and $ h $ is the input embedding dimension. The Multi-Head Attention (MHA) block computes a set of queries ($ Q $), keys ($ K $), and values ($ V $) through linear projections of $ X^i $:

\begin{equation}
Q^i = X^i \cdot W^i_Q, \quad K^i = X^i \cdot W^i_K, \quad V^i = X^i \cdot W^i_V,
\label{eq:mha_projection}
\end{equation}
where $ W^i_Q, W^i_K, W^i_V \in \mathbb{R}^{h \times h} $ are the projection matrices. The generated $K^i$ and $V^i$ are stored in the KV cache.

\noindent The self-attention score in MHA is computed as:

\begin{equation}
Z^i = \text{softmax}\left(\frac{Q^i (K^i)^T}{\sqrt{d_{\text{head}}}}\right) \cdot V^i,
\label{eq:attention}
\end{equation}
where $ d_{\text{head}} $ represents the dimension of each attention head. 
Finally, the attention score is applied with a linear projection to produce the output of the MHA block:

\begin{equation}
O^i = Z^i \cdot W^i_O,
\label{eq:mha_output}
\end{equation}
where $ W^i_O \in \mathbb{R}^{h \times h} $ is the projection matrix.

The feedforward network (FFN) is followed after the MHA block, which consists of two fully connected layers with a non-linear activation function applied between them. 
It processes the attention output $ O^i $ to generate the input for the next decoder layer as follows:

\begin{equation}
X^{i+1} = \sigma(O^i \cdot W^i_1) \cdot W^i_2,
\label{eq:ffn}
\end{equation}
where $ W^i_1 \in \mathbb{R}^{h \times d_{\text{FFN}}} $ and $ W^i_2 \in \mathbb{R}^{d_{\text{FFN}} \times h} $ are the weight matrices of the two linear layers, and $ \sigma(\cdot) $ denotes the activation function.

In the decoding stage, the $i$-th decoder layer receives a single token $ x^i \in \mathbb{R}^{b \times 1 \times h} $. The KV cache is updated by concatenating the newly computed key and value pairs with the existing ones:


\begin{equation}
\begin{aligned}
K^i &= \text{concat}(K^i, x^i \cdot W^i_K), \\
V^i &= \text{concat}(V^i, x^i \cdot W^i_V).
\end{aligned}
\label{eq:kv_update}
\end{equation}

The remaining attention and feedforward computations in the decoding stage are identical to those in the prefilling stage.

\section{Proposed Method}
\textbf{LLM inference scheduling.} Our approach aims at LLM inference systems with large KV caches that are stored on CPU DRAM and fetched into GPU memory as needed. Since LLMs have many layers and many batches of inputs to process, there are different scheduling strategies to determine how computations are performed across batches and layers to optimize for specific performance goals, such as minimizing latency or maximizing throughput.  
Row-by-row schedule (as shown in Appendix \ref{app:schedule}) processes one batch at a time, using layer-wise execution. In this scenario, model weights are kept in GPU memory whenever feasible. If the model weights are also offloaded to the CPU, both the KV cache and the model weights for a single layer are transferred to the GPU, processed for the current batch, and then cleared. This process is repeated layer by layer until a token is generated. When minimizing latency is the primary goal, this approach is preferred because all prompts in a batch are fully processed to generate their complete context before proceeding to the next batch.

Column-by-column scheduling (Appendix \ref{app:schedule}) is more effective for maximizing throughput by increasing the \emph{effective batch size} (number of batches times batch size) to process more sequences in parallel, at the cost of longer latency. In this approach, the model weights are offloaded to CPU memory to accommodate a large batch size. The model weights and KV cache for a single layer are transferred to GPU memory and processed for the first batch. Instead of moving to the next layer for the current batch, subsequent batches are processed using the same layer while keeping the weights stationary in GPU memory. Once a group of batches is processed for the first layer, the process moves to the second layer for each batch. Note that the effective batch size is limited by the available storage for activations and KV cache, as they must still be stored in CPU memory or external storage once they exceed the GPU memory capacity. 

Our proposed design is independent of the scheduling strategy, whether row-by-row or column-by-column, and aims to overlap the majority of the PCIe transfer time with GPU computations, thereby improving overall efficiency.

\subsection{Design Overview}

\begin{figure*}
    \centering
    \includegraphics[width=0.7\textwidth]{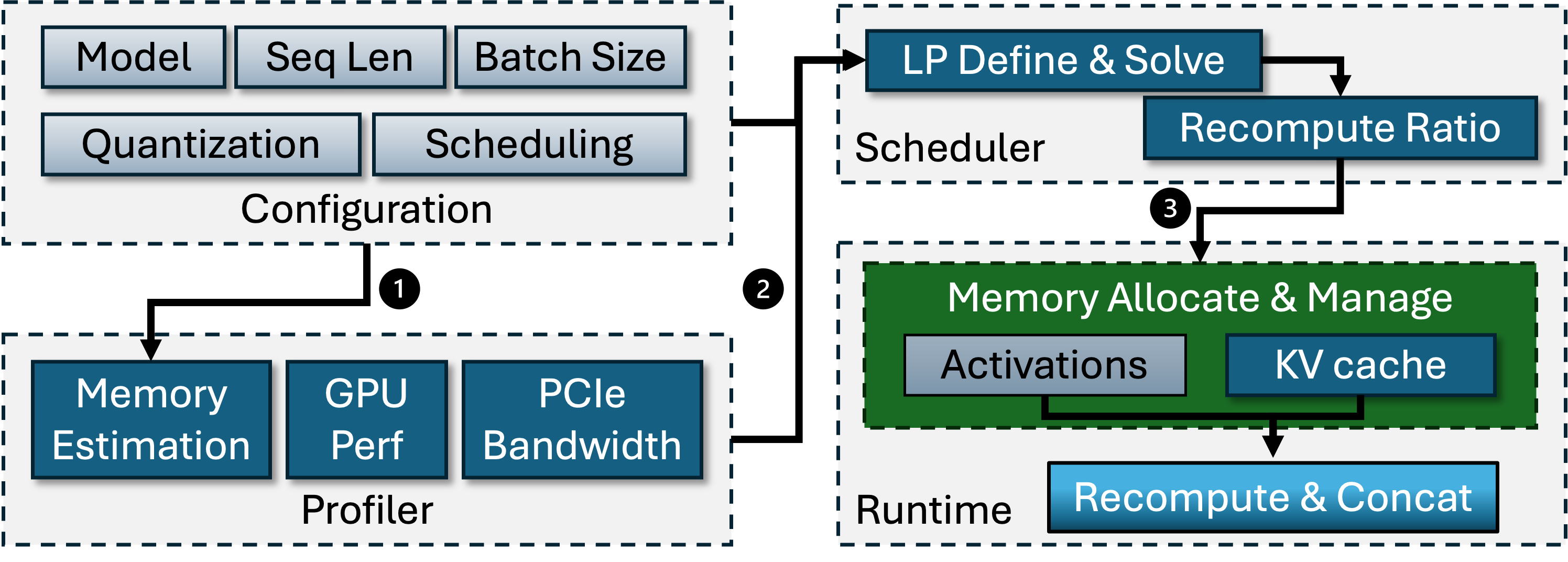}
    \caption{Design overview of {\paperName}. User configuration and profiling inform the scheduler, which computes an optimal KV cache recompute ratio. The runtime then overlaps data transfer and GPU computation to improve inference efficiency.}
    \label{fig:design-overview}
\end{figure*}

\noindent To relieve PCIe pressure and improve GPU computation utilization, we propose a novel method, {\paperName}, that recomputes partial KV cache on the GPU while transferring the rest of the KV cache to the GPU. As shown in Figure \ref{fig:design-overview}, {\paperName} comprises three main modules: the profiler, scheduler, and runtime. 
User configuration includes performance objective (i.e., latency or throughput), data parameters such as prompt length, generation length, batch size, and model information like input embedding dimension and number of layers. 
The \textbf{profiler module} gathers system statistics, which provide insights into hardware characteristics like PCIe bandwidth and GPU processing speed. For example, the profiler module utilizes the batch size, model information, and sequence length to characterize PCIe bandwidth. Using this information along with the user configuration, the \textbf{scheduler module} calculates the best KV cache split point for recomputation by solving a linear programming problem, aiming to maximize the overlap between the computation and communication operations and utilization of both GPU and PCIe bandwidth during the inference process. The \textbf{runtime module}, in turn, utilizes this execution strategy to process user inputs and manage the memory allocation and data transfer streams. 

\subsection{Scheduler Module}
In this section, we describe how {\paperName} is adopted to either the row-by-row or column-by-column schedule. 

\begin{figure}[h]
\centering
\begin{subfigure}[b]{\columnwidth}
    \centering
    \includegraphics[width=\columnwidth]{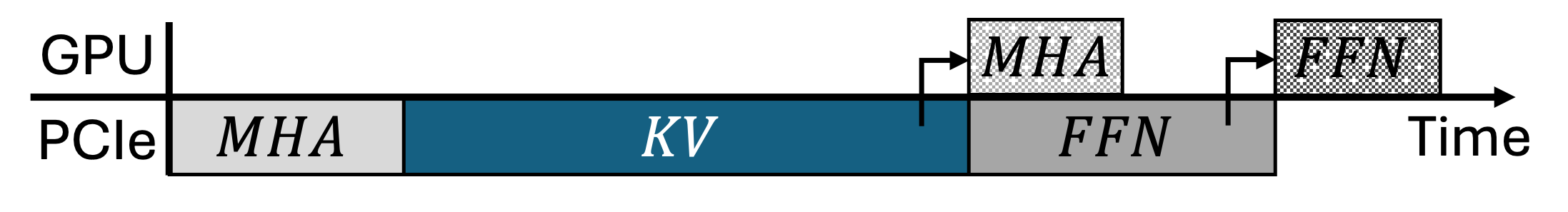}
    \caption{Naive offloading pipeline for row-wise scheduling with asynchronous data transfer. GPU and PCIe denote GPU computation and data transfer, with arrows indicating data dependencies.}
\end{subfigure}

\begin{subfigure}[b]{\columnwidth}
    \centering
    \includegraphics[width=\columnwidth]{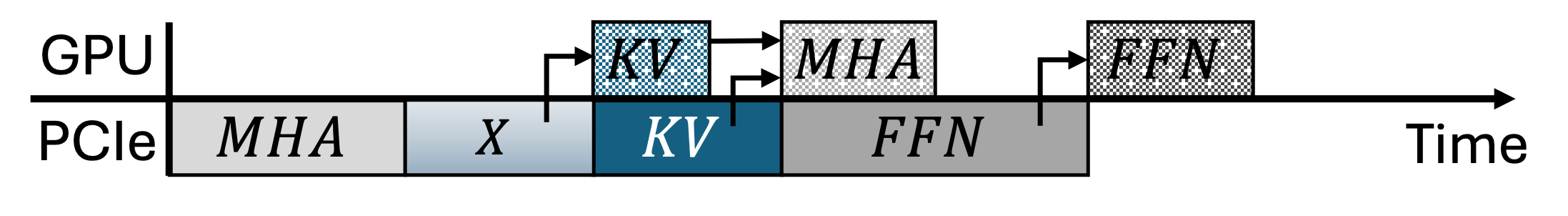}
    \caption{Offloading pipeline for row-wise scheduling with KV cache partial recomputation.}
\end{subfigure}

\caption{Comparison of two offloading pipelines.}
\label{fig:row-by-row}
\end{figure}

\noindent \textbf{Row-by-row schedule with KV cache partial recomputation.}
If the performance objective is to minimize latency, the scheduler module will initiate a row-by-row execution plan. The naive offloading pipeline of a row-by-row schedule is shown in Figure \ref{fig:row-by-row}(a), where both the KV cache and model weights are offloaded to CPU memory. The required data are transferred asynchronously over PCIe to the GPU for executing the MHA and FFN blocks. Storing newly generated KV pairs to CPU memory is omitted from the figure for simplicity. 
Since the KV cache is larger in size compared to the MHA weights, it arrives at the GPU later during the asynchronous transfer. The pipeline is slightly different if model weights are not offloaded to the CPU. In this case, only the MHA block will wait for the KV cache data to be transferred to the GPU before starting the computation.

In {\paperName}, rather than transferring the entire KV cache from CPU memory to GPU memory, the GPU recomputes partial KV cache using corresponding input activations that are transferred from CPU first, while the remaining KV cache is asynchronously transferred to the GPU, as illustrated in Figure \ref{fig:row-by-row}(b). The GPU then merges the recomputed KV cache with the transferred KV cache to perform MHA computations.

\begin{figure}[h]
\centering
    \includegraphics[width=\columnwidth]{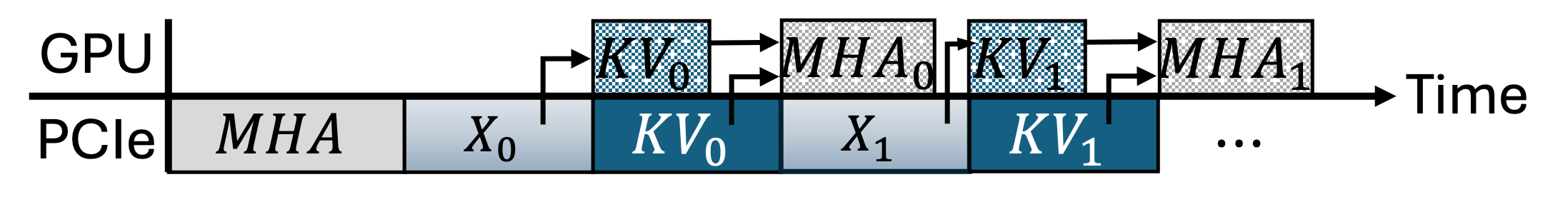}
\caption{Offloading pipeline for column-wise scheduling with KV cache partial recomputation to maximize throughput.}
\label{fig:column-by-column-recomputation}
\end{figure}

\noindent \textbf{Column-by-column schedule with KV cache partial recomputation.}
When the performance objective is to maximize throughput, the scheduler module adopts a column-by-column execution plan. This approach, illustrated in Figure \ref{fig:column-by-column-recomputation}, accommodates large batch size inference by reusing model weights across multiple batches.
As soon as the KV cache for batch 0 is fully transmitted, the activations for batch 1 are transferred. Simultaneously, the GPU begins computing the MHA for batch 0. Unlike the row-by-row schedule, which processes all layers sequentially within a single batch before moving to the next batch, the column-by-column schedule processes multiple batches on the same layer. As a result, activations corresponding to the recomputed KV cache must be stored until generation for that batch is complete.

\noindent \textbf{Determining the optimal KV cache split point.} 
In both scheduling methods, the objective is to identify the optimal split point, which defines the division of the KV cache between the portion recomputed on the GPU and the portion transferred from CPU memory. This problem can be formulated as a linear programming problem.
The row-by-row schedule can be viewed as a special case of the column-by-column schedule, where activations for recomputing the KV cache are not transferred. We first formulate the problem for the column-by-column schedule and then demonstrate how it simplifies to the row-by-row schedule.

Given the current sequence length $s'$, which is greater than the prompt length $s$, the activation transferred to the GPU in the $i$-th layer is represented by $ {X}^i[0:l] $, where $0 \leq l \leq s'$. The remaining KV cache for the subsequent tokens is denoted by $ {K}^i[l:s'] $ and $ {V}^i[l:s'] $. The memory usage of these activations is:

\begin{equation}
\begin{aligned}
M_{{X}^i[0:l]} &= b \times l \times h \times p, \\
{M}_{{KV}^i[l:s']} &= 2 \times b \times (s' - l) \times h \times p.
\end{aligned}
\end{equation}

\noindent Recomputing the KV cache for $ {X}^i[0:l] $ requires:  

\begin{equation}
\begin{aligned}
{K}^i[0:l] &= {X}^i[0:l] \cdot {W}^i_K, \\
{V}^i[0:l] &= {X}^i[0:l] \cdot {W}^i_V.
\end{aligned}
\label{eq:recompute}
\end{equation}

\noindent This recomputation on the GPU requires floating-point operations of

\begin{equation}
N_{{KV}^i[0:l]} = 4 \times b \times l \times h^2.
\end{equation}

\noindent Consequently, the recomputation time $ t_{gpu}^i $ for the KV cache is given by

\begin{equation}
t_{recomp}^i = \frac{N_{{KV}^i[0:l]}}{v_{gpu}},
\end{equation}

\noindent where $ v_{gpu} $ denotes the GPU processing speed. The total time $ t^i $ for processing is as follows: 

\begin{equation}
t^i = \frac{{M}_{{X}^i[0:l]}}{v_{com}} + \max\left(t_{recomp}^i, \frac{{M}_{{KV}^i[l:s']}}{v_{com}}\right),
\label{eq:lp_target}
\end{equation}

\noindent where $ v_{com} $ represents the data transmission speed for activations and KV cache.

The objective is to determine the optimal $ l $ that minimizes this total processing time $ t^i $, which becomes a linear programming problem:  

\begin{equation}
\begin{aligned}
\min_{l} \quad & t^i \\
\text{s.t.} \quad & 0 \leq l \leq s \quad \forall i \in \{1, \ldots, n\}.
\end{aligned}
\label{eq:lp}
\end{equation}

The optimal split point $l$ depends on the current sequence length $s'$, which increases during generation and must therefore be determined adaptively. Fortunately, solving this linear programming problem is computationally negligible because there is only one integer variable. If the first term in Eq.~\eqref{eq:lp_target} is omitted, the problem simplifies to the row-by-row schedule.

\subsection{Runtime Module} 

\textbf{Asynchronous overlapping.}
To enable concurrent execution of GPU computation and CPU-GPU communication, the runtime module employs a communication parallelism strategy with six processes: weight loading, KV cache loading, activation loading, recomputed activation loading, KV cache storing, and activation storing, as detailed in Appendix \ref{app:overlapping}. By incorporating double buffering and prefetching techniques, it simultaneously loads weights for the next layer, and retrieves activations for KV cache recomputation and KV cache for the next batch, while storing cache and activations from the previous batch and processing the current batch.

\noindent \textbf{Pinned memory.}
To optimize data transfer, like prior works \cite{sheng2023flexgen, yu2024twinpilots}, we utilize pinned CPU memory for recomputed activation and the weights that are transferred to the GPU. Using pinned memory enables faster and asynchronous transfer, as it avoids the need to page data in and out.

\begin{figure}[h]
\centering
\begin{subfigure}[t]{\columnwidth}
\centering
    \includegraphics[width=\columnwidth]{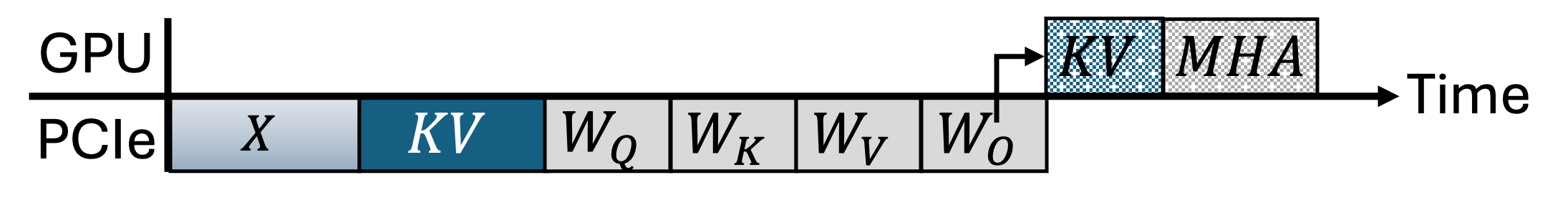}
\caption{Coarse-grained offloading pipeline with delayed KV cache partial recomputation.}
\end{subfigure}

\begin{subfigure}[t]{\columnwidth}
\centering
    \includegraphics[width=\columnwidth]{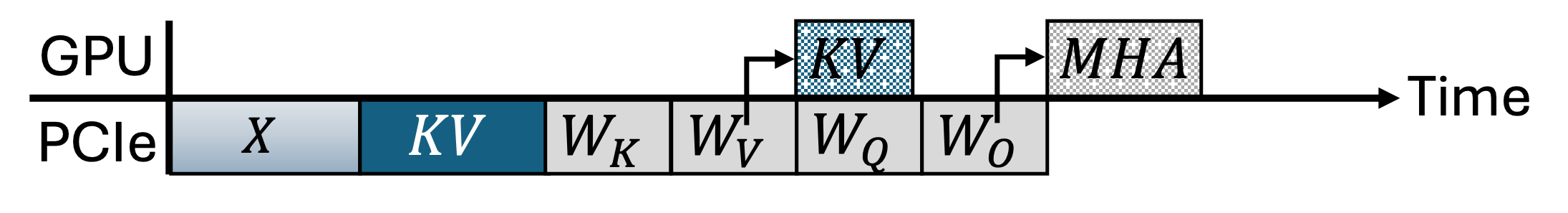}
\caption{Fine-grained offloading pipeline overlapping KV cache recomputation with weights loading.}
\end{subfigure}

\caption{Comparison of offloading pipelines with different levels of granularity in the MHA layer.}
\label{fig:hiding-recomputation-pipeline}
\end{figure}

\noindent \textbf{Hiding KV cache partial recomputation.}
If both the KV cache and model weights are offloaded, and the size of the transferred KV cache is smaller than the size of the model weights, a coarse-grained computation pipeline with KV cache partial recomputation may degrade inference performance. This occurs because recomputation waits until all MHA weights ($W_Q$, $W_K$, $W_V$, and $W_O$) are fully loaded, as shown in Figure \ref{fig:hiding-recomputation-pipeline}(a), which delays the MHA computation. However, KV cache recomputation only requires $W_K$ and $W_V$ (Eq. ~\eqref{eq:recompute}), making it unnecessary to wait for the complete weight loading process. To address this, we implement a fine-grained MHA pipeline that prioritizes loading $W_K$ and $W_V$ first. Once these weights are available, KV cache recomputation can begin immediately. As illustrated in Figure \ref{fig:hiding-recomputation-pipeline}(b), $W_K$ and $W_V$ are used for KV cache partial recomputation, followed by the use of $W_Q$ and $W_O$ for MHA computation. This approach effectively overlaps KV cache recomputation with weight loading, ensuring that in the worst-case scenario, the method performs no worse than the baseline bottlenecked by weight loading.

\section{Experiments}
\label{sec:exp}
\textbf{Hardware.}
In our experiments, we utilize an NVIDIA A100 GPU with 40 GB of memory, connected to the CPU through a PCIe 4.0 x16 interface, which provides a bandwidth of 32 GB/s. The CPU is an AMD EPYC processor with 64 cores, operating at 2.6 GHz.
Our method and implementation automatically adapt to the underlying hardware, which allows for flexible deployment across diverse system architectures.

\noindent \textbf{Model.}
We evaluate {\paperName} using OPT models \cite{zhang2022opt} with parameter sizes ranging from 6.7B to 30B. While our experiments focus on OPT models, the recomputation technique presented in this work is compatible with other LLM architectures, such as LLaMa \cite{touvron2023llama} and GPT-3 \cite{brown2020language}, due to their similar attention mechanisms \cite{ashish2017attention}.

\noindent \textbf{Workload.}
We evaluate {\paperName} on two types of workloads: latency-oriented and throughput-oriented. In the latency-oriented workload, the model weights are retained in GPU memory to avoid the costly repeated loading. Due to the limited memory size of a  GPU, experiments are conducted using OPT-6.7B and OPT-13B. In the throughput-oriented workload, model weights are offloaded to the CPU after computation to free more GPU memory for handling larger batches. This setup is evaluated using OPT-6.7B, OPT-13B, and OPT-30B.

To provide accurate comparisons, we use the same datasets as those in FlexGen \cite{sheng2023flexgen} with prompts uniformly padded to the same length, with models configured to generate 32 or 128 tokens per prompt. To evaluate performance across different input scenarios, our evaluation uses prompt lengths of 256, 512, and 1024 tokens. Performance metrics include decoding latency (time taken to generate tokens) for latency-oriented workloads and decoding throughput (tokens generated per second) for throughput-oriented workloads, as {\paperName} does not impact prefilling performance. We report an average decoding latency and throughput across five test runs, respectively.

\noindent \textbf{Baseline.}
In our experiments, we use DeepSpeed Inference \cite{aminabadi2022deepspeed}, Hugging Face Accelerate \cite{accelerate} as the baseline for latency-oriented workload experiments, as Hugging Face Transformers library currently supports KV cache offloading to CPU memory while still retaining the model weights in GPU memory. We use FlexGen \cite{sheng2023flexgen} as the baseline for throughput-oriented workload experiments, as it supports column-by-column schedule by offloading both model weights and KV cache to the CPU.

\noindent \textbf{Implementation.} {\paperName} is implemented on top of Hugging Face Transformers (v4.46.1) \cite{wolf2020transformers} and FlexGen \cite{sheng2023flexgen} frameworks to ensure fair comparison with baselines. In the Transformers implementation, we utilize double buffering in GPU memory to overlap KV cache transfer across decoder layers. For both the Transformers and FlexGen implementations, we utilize CUDA streams to enable asynchronous overlapping as described in Algorithm \ref{alg:block_schedule_overlap}.

\subsection{Latency-oriented Experiments}

\begin{figure*}[t]
\centering
    \begin{subfigure}[t]{0.32\textwidth}
        \centering
        \begin{tikzpicture}
            \begin{axis}[
                axis line style={line width=1pt},
                title={OPT-6.7B},
                title style={yshift=-5pt},
                width=1.1\textwidth,
                height=4cm,
                ybar=0pt,
                bar width=.2cm,
                tick label style={font=\scriptsize},
                symbolic x coords={A, B, C, D, E, F},
                xticklabel style={yshift=5pt, align=center},
                xticklabels={{256 \\ 32}, {256 \\ 128}, {512 \\ 32}, {512 \\ 128}, {1024 \\ 32}, {1024 \\ 128}},
                xtick=data,
                xtick style={draw=none},
                ytick={0, 20, 40, 60, 80},
                ymin=0,
                xlabel={Sequence length},
                ylabel={Throughput (tokens/s)},
                ylabel style={yshift=-10pt},
                enlarge x limits=0.15,
                grid=major,
                grid style=solid,
                major x grid style={draw=none},
                legend entries={FlexGen, {\paperName}},
                legend style={at={(1,1)}, anchor=north east, font=\scriptsize, row sep=0cm, column sep=0cm, legend cell align=left},
                legend image code/.code={\draw[yshift=-0.06cm] rectangle (0.2cm,0.2cm);},
            ]
            \addplot[fill=gray!] coordinates {(A, 62.04) (B, 62.153) (C, 38.307) (D, 39.823) (E, 22.509) (F, 23.886)};
            \addplot[fill=darkteal] coordinates {(A, 84.316) (B, 77.863) (C, 48.213) (D, 48.283) (E, 25.543) (F, 27.396)};
            \end{axis}
        \end{tikzpicture}
    \end{subfigure}
    \begin{subfigure}[t]{0.32\textwidth}
        \centering
        \begin{tikzpicture}
            \begin{axis}[
                axis line style={line width=1pt},
                title={OPT-13B},
                title style={yshift=-5pt},
                width=1.1\textwidth,
                height=4cm,
                ybar=0pt,
                bar width=.2cm,
                tick label style={font=\scriptsize},
                symbolic x coords={A, B, C, D, E, F},
                xticklabel style={yshift=5pt, align=center},
                xticklabels={{256 \\ 32}, {256 \\ 128}, {512 \\ 32}, {512 \\ 128}, {1024 \\ 32}, {1024 \\ 128}},
                xtick=data,
                xtick style={draw=none},
                ytick={0, 10, 20, 30, 40, 50},
                ymin=0,
                xlabel={Sequence length},
                enlarge x limits=0.15,
                grid=major,
                grid style=solid,
                major x grid style={draw=none},
            ]
            \addplot[fill=gray!] coordinates {(A, 39.096) (B, 38.339) (C, 24.659) (D, 23.832) (E, 13.582) (F, 14.997)};
            \addplot[fill=darkteal] coordinates {(A, 49.418) (B, 44.022) (C, 33.054) (D, 30.76) (E, 19.817) (F, 18.364)};
            \end{axis}
        \end{tikzpicture}
    \end{subfigure}
    \begin{subfigure}[t]{0.32\textwidth}
        \centering
        \begin{tikzpicture}
            \begin{axis}[
                axis line style={line width=1pt},
                title={OPT-30B},
                title style={yshift=-5pt},
                width=1.1\textwidth,
                height=4cm,
                ybar=0pt,
                bar width=.2cm,
                tick label style={font=\scriptsize},
                symbolic x coords={A, B, C, D, E, F},
                xticklabel style={yshift=5pt, align=center},
                xticklabels={{256 \\ 32}, {256 \\ 128}, {512 \\ 32}, {512 \\ 128}, {1024 \\ 32}, {1024 \\ 128}},
                xtick=data,
                xtick style={draw=none},
                ytick={0, 5, 10, 15, 20, 25},
                ymin=0,
                xlabel={Sequence length},
                ylabel style={at={(1.4,-0.3)}, anchor=west},
                ylabel={Effective bs = 32 x 8},
                enlarge x limits=0.15,
                grid=major,
                grid style=solid,
                major x grid style={draw=none},
            ]
            \addplot[fill=gray!] coordinates {(A, 21.912) (B, 21.108) (C, 13.096) (D, 13.479) (E, 7.61) (F, 8.127)};
            \addplot[fill=darkteal] coordinates {(A, 25.064) (B, 23.692) (C, 16.829) (D, 16.917) (E, 9.456) (F, 10.485)};
            \end{axis}
        \end{tikzpicture}
    \end{subfigure}
    \begin{subfigure}[t]{0.32\textwidth}
        \centering
        \begin{tikzpicture}
            \begin{axis}[
                axis line style={line width=1pt},
                width=1.1\textwidth,
                height=4cm,
                ybar=0pt,
                bar width=.2cm,
                tick label style={font=\scriptsize},
                symbolic x coords={A, B, C, D, E, F, G},
                xticklabel style={yshift=5pt, align=center},
                xticklabels={1, 2, 4, 8, 16, 32, 48},
                xtick=data,
                xtick style={draw=none},
                ytick={0, 5, 10, 15, 20, 25},
                ymin=0,
                xlabel={Batch size},
                xlabel style={yshift=5pt},
                ylabel={Throughput (tokens/s)},
                ylabel style={yshift=-10pt},
                enlarge x limits=0.15,
                grid=major,
                grid style=solid,
                major x grid style={draw=none},
            ]
            \addplot[fill=gray!] coordinates {(A, 8.799) (B, 15.393) (C, 17.89) (D, 18.075) (E, 19.521) (F, 22.509) (G, 20.435)};
            \addplot[fill=darkteal] coordinates {(A, 9.046) (B, 16.573) (C, 20.895) (D, 23.941) (E, 25.712) (F, 25.543) (G, 26.241)};
            \end{axis}
        \end{tikzpicture}
    \end{subfigure}
    \begin{subfigure}[t]{0.32\textwidth}
        \centering
        \begin{tikzpicture}
            \begin{axis}[
                axis line style={line width=1pt},
                width=1.1\textwidth,
                height=4cm,
                ybar=0pt,
                bar width=.2cm,
                tick label style={font=\scriptsize},
                symbolic x coords={A, B, C, D, E, F, G},
                xticklabel style={yshift=5pt, align=center},
                xticklabels={1, 2, 4, 8, 16, 32, 48},
                xtick=data,
                xtick style={draw=none},
                ytick={0, 5, 10, 15, 20},
                ymin=0,
                xlabel={Batch size},
                xlabel style={yshift=5pt},
                ylabel style={yshift=-10pt},
                enlarge x limits=0.15,
                grid=major,
                grid style=solid,
                major x grid style={draw=none},
            ]
            \addplot[fill=gray!] coordinates {(A, 3.827) (B, 6.263) (C, 8.211) (D, 10.455) (E, 11.507) (F, 12.313) (G, 15.027)};
            \addplot[fill=darkteal] coordinates {(A, 3.821) (B, 6.908) (C, 9.835) (D, 12.621) (E, 15.248) (F, 18.007) (G, 17.55)};
            \end{axis}
        \end{tikzpicture}
    \end{subfigure}
    \begin{subfigure}[t]{0.32\textwidth}
        \centering
        \begin{tikzpicture}
            \begin{axis}[
                axis line style={line width=1pt},
                width=1.1\textwidth,
                height=4cm,
                ybar=0pt,
                bar width=.2cm,
                tick label style={font=\scriptsize},
                symbolic x coords={A, B, C, D, E, F, G},
                xticklabel style={yshift=5pt, align=center},
                xticklabels={1, 2, 4, 8, 16, 32, 48},
                xtick=data,
                xtick style={draw=none},
                ytick={0, 2.5, 5, 7.5, 10},
                ymin=0,
                xlabel={Batch size},
                xlabel style={yshift=5pt},
                ylabel style={at={(1.4,-0.3)}, anchor=west},
                ylabel={Seq length=1024+32},
                enlarge x limits=0.15,
                grid=major,
                grid style=solid,
                major x grid style={draw=none},
            ]
            \addplot[fill=gray!] coordinates {(A, 1.989) (B, 3.1) (C, 4.56) (D, 5.935) (E, 6.785) (F, 7.61) (G, 8.172)};
            \addplot[fill=darkteal] coordinates {(A, 2.012) (B, 3.318) (C, 5.009) (D, 6.544) (E, 8.567) (F, 9.456) (G, 10.613)};
            \end{axis}
        \end{tikzpicture}
    \end{subfigure}

    \caption{Throughput comparison for various models and configurations.}
    \label{fig:throughput_exp}
\end{figure*}

We evaluate the decoding latency required to complete a single batch for settings of different sequence lengths. Figure \ref{fig:latency-exp} shows that {\paperName} consistently outperforms the baselines, DeepSpeed Inference and Hugging Face Accelerate, for both OPT-6.7B and OPT-13B. 
The experimental results show that {\paperName} reduces decoding latency, especially at longer generation lengths. For instance, OPT 6.7B at a prompt length of 128 with 128 tokens generated, the latency is reduced by approximately 35.8\% compared to Hugging Face Accelerate. Detailed experiential results including KV cache size, GPU peak memory usage, and optimal recomputation split points over the generation process are provided in Appendix \ref{app:detailed-exp} and \ref{app:optimal-split-points}. 

\begin{figure}[h]
    \centering
    \begin{tikzpicture}
        \begin{axis}[
            axis line style={line width=1pt},
            width=\columnwidth,
            height=5cm,
            ybar=0pt,
            bar width=.15cm,
            tick label style={font=\scriptsize},
            symbolic x coords={A, B, C, D, E, F},
            xticklabel style={align=center},
            xticklabels={{128 \\ 32}, {128 \\ 128}, {256 \\ 32}, {256 \\ 128}, {512 \\ 32}, {512 \\ 128}},
            xtick=data,
            xtick style={draw=none},
            xticklabel style={yshift=5pt, align=center},
            xlabel={Sequence length},
            ylabel={Latency (seconds)},
            ylabel style={yshift=-10pt},
            ymin=0,
            ytick={0, 25, 50, 75, 100, 125, 150, 175},
            grid=major,
            grid style=solid,
            major x grid style={draw=none},
            legend entries={DS (OPT-6.7B), Accel. (OPT-6.7B), {\paperName} (OPT-6.7B), DS (OPT-13B), Accel. (OPT-13B), {\paperName} (OPT-13B)},
            legend style={at={(0,1)}, anchor=north west, font=\scriptsize, row sep=0cm, column sep=0cm, legend cell align=left, legend columns=2},
            legend image code/.code={\draw[yshift=-0.06cm] rectangle (0.2cm,0.2cm);},
        ]

        \addplot[fill=black] coordinates {(A,9.613) (B,74.805) (C,29.892) (D,93.965) (E,25.162) (F,112.516)};

        \addplot[fill=gray] coordinates {(A,8.905) (B,71.327) (C,26.825) (D,88.354) (E,24.390) (F,110.277)};

        \addplot[fill=darkteal] coordinates {(A,6.651) (B,45.766) (C,19.138) (D,61.597) (E,20.349) (F,93.932)};

        \addplot[pattern=north west lines, pattern color=black] coordinates {(A,11.525) (B,74.015) (C,22.740) (D,119.806) (E,39.956) (F,189.124)};

        \addplot[pattern=north east lines, pattern color=gray] coordinates {(A,11.409) (B,73.896) (C,19.381) (D,104.115) (E,35.066) (F,168.155)};

        \addplot[pattern=north west lines, pattern color=darkteal] coordinates {(A,9.148) (B,66.119) (C,16.654) (D,88.492) (E,29.215) (F,138.377)};

        \end{axis}
    \end{tikzpicture}
    \caption{Decoding latency for a single batch of size 64 across different sequence lengths.}
    \label{fig:latency-exp}
\end{figure}

\subsection{Throughput-oriented Experiments}

We also evaluate throughput performance during the decoding stage, as {\paperName} does not affect the prefilling stage. To maximize throughput, we set the effective batch size to be 32 by 8, meaning each layer computes on 8 batches of size 32 sequentially before moving to the next layer. The first row of Figure \ref{fig:throughput_exp} shows the results, demonstrating that {\paperName} consistently outperforms FlexGen under settings of all sequence lengths for different models. It achieves up to 15.1\%, 46.2\%, and 29.0\% speedup in throughput for OPT-6.7B, OPT-13B, and OPT-30B, respectively. Additional experimental results on a low-end GPU system are provided in Appendix \ref{app:low-end-gpu-experiments}.

We also compare {\paperName} with FlexGen for varying batch sizes from 1 to 48 with a fixed prompt length of 1,024 and a generation length of 32, as shown in the second row of Figure \ref{fig:throughput_exp}. {\paperName} consistently outperforms FlexGen across all batch sizes. As the KV cache grows larger, {\paperName} shows greater performance benefits due to reduced KV cache transfer over the PCIe bus.

\subsection{GPU Utilization}

\noindent To evaluate the efficiency improvement, we analyze the temporal resource utilization of {\paperName} and FlexGen as shown in Figure \ref{fig:ablation-utilization}. At first in the prefilling stage, both methods reach full GPU utilization since the prefilling stage is compute-bound. However, in the decoding stage, in contrast to FlexGen, {\paperName} enhances GPU utilization, increasing it from 85\% to 99\% on average by overlapping GPU computations with CPU-GPU data transfer, while maintaining the same peak memory usage indicated by the black lines.

\begin{figure}[h]
\centering
\begin{tikzpicture}
    \begin{axis}[
        axis line style={line width=1pt},
        width=0.95\columnwidth,
        height=4cm,
        tick label style={font=\scriptsize},
        xlabel={Time (s)},
        xlabel style={yshift=5pt},
        ylabel={Percentage},
        ylabel style={yshift=-10pt},
        ymin=0, ymax=110,
        xtick={0,40,80,120,160,200,240},
        ytick={0,25,50,75,100},
        grid=major,
        legend entries={{\paperName} (C), FlexGen (C), {\paperName} (M), FlexGen (M)},
        legend style={at={(0.5, 1)}, anchor=south, font=\scriptsize, column sep=0cm, legend cell align=left, legend columns=4},
        legend image post style={xscale=0.26}
    ]

        \addplot[solid, color=darkteal, thick] coordinates {
            (0, 0) (10, 0) (20, 100) (30, 100) (40, 100) (50, 100) 
            (60, 100) (70, 100) (80, 77) (90, 89) (100, 99) (110, 99) 
            (120, 99) (130, 86) (140, 99) (150, 99) (160, 99) (170, 99) 
            (180, 0) (190, 0)
        };
        
        \addplot[dotted, color=darkteal, very thick] coordinates {
            (0, 0) (10, 0) (20, 98) (30, 100) (40, 100) (50, 100) 
            (60, 100) (70, 100) (80, 100) (90, 85) (100, 87) (110, 74) 
            (120, 85) (130, 85) (140, 86) (150, 89) (160, 73) (170, 80) 
            (180, 85) (190, 70) (200, 82) (210, 85) (220, 86) (230, 0) 
            (240, 0)
        };

        \addplot[solid, color=black, thick] coordinates {
            (0, 0) (10, 31) (20, 57.5) (30, 57.5) (40, 57.5) (50, 57.5) 
            (60, 57.5) (70, 57.5) (80, 70) (90, 70) (100, 70) (110, 70) 
            (120, 70) (130, 70) (140, 70) (150, 70) (160, 70) (170, 72.5) 
            (180, 72.5) (190, 0)
        };

        \addplot[dotted, color=black, very thick] coordinates {
            (0, 0) (10, 25) (20, 41) (30, 52) (40, 48) (50, 55) 
            (60, 55) (70, 55) (80, 54) (90, 70) (100, 70) (110, 70) 
            (120, 70) (130, 70) (140, 70) (150, 70) (160, 70) (170, 70) 
            (180, 70) (190, 70) (200, 70) (210, 72.5) (220, 72.5) 
            (230, 72.5) (240, 0)
        };

    \end{axis}
\end{tikzpicture}
\caption{Computation and memory resource usage of {\paperName} and FlexGen during decoding stage.}
\label{fig:ablation-utilization}
\end{figure}

\subsection{KV Cache Compression}

We apply group-wise 4-bit quantization to compress the KV cache, which has been shown to have minimal impact on model accuracy \cite{sheng2023flexgen}. Figure \ref{fig:compression} shows that applying compression reduces the amount of data transferred to the GPU, leading to further improvements in decoding throughput. These results showcase the compatibility of {\paperName} with KV cache compression and its potential to achieve additional performance gains by alleviating PCIe bandwidth bottlenecks.

\begin{figure}[h]
\centering
    \begin{tikzpicture}
        \begin{axis}[
            axis line style={line width=1pt},
            width=0.95\columnwidth,
            height=4.3cm,
            ybar=0pt,
            bar width=.2cm,
            tick label style={font=\scriptsize},
            symbolic x coords={A, B, C, D, E, F},
            xticklabel style={align=center},
            xticklabels={{256 \\ 32}, {256 \\ 128}, {512 \\ 32}, {512 \\ 128}, {1024 \\ 32}, {1024 \\ 128}},
            xtick=data,
            xtick style={draw=none},
            xticklabel style={yshift=5pt, align=center},
            xlabel={Sequence length},
            ylabel={Throughput (tokens/s)},
            ylabel style={yshift=-10pt},
            ymin=0,
            enlarge x limits=0.15,
            grid=major,
            grid style=solid,
            major x grid style={draw=none},
            legend entries={w/o KV cache compression, w/ KV cache compression},
            legend style={at={(1,1)}, anchor=north east, font=\scriptsize, row sep=0cm, column sep=0cm, legend cell align=left},
            legend image code/.code={\draw[yshift=-0.06cm] rectangle (0.2cm,0.2cm);},
        ]
    \addplot[fill=gray] coordinates {(A, 49.418) (B, 44.022) (C, 33.054) (D, 30.76) (E, 19.817) (F, 18.364)};
    \addplot[fill=darkteal] coordinates {(A, 65.423) (B, 63.098) (C, 45.054) (D, 43.502) (E, 26.537) (F, 25.969)};
    \end{axis}
    \end{tikzpicture}
\caption{Decoding throughput improvement with KV cache compression enabled on OPT-13B model.}
\label{fig:compression}
\end{figure}

\subsection{Ablation Study}

\textbf{Hiding KV cache partial recomputation.}
To evaluate the effectiveness of the fine-grained offloading pipeline that overlaps KV cache recomputation with weight loading, we conduct experiments using the OPT-6.7B model. In this ablation, we use a small KV cache size to ensure that MHA weights always arrive at the GPU later than the KV cache.
Table \ref{tab:ablation-hiding} presents decoding latency across varying smaller batch sizes, comparing three configurations: FlexGen, {\paperName} without hiding KV cache recomputation, and {\paperName} with hiding. When the batch size is 1 and the KV cache size is the smallest, FlexGen can outperform {\paperName} without hiding.
By overlapping the transfer of MHA weights with KV cache recomputation, {\paperName} ensures performance that is no worse than FlexGen under this scenario, particularly when weight loading is the primary bottleneck. This result shows that {\paperName} works well for both small and large batch size settings, thereby providing a unified approach to improve decoding performance.

\begin{table}[h]
\centering
\resizebox{\columnwidth}{!}{%
\begin{tabular}{lcccccc}
\hline
\textbf{Batch size} & \textbf{1} & \textbf{2} & \textbf{4} & \textbf{8} & \textbf{16} & \textbf{32} \\ \hline
KV cache (MB) & 3 & 6 & 12 & 24  & 48  & 64 \\ \hline
FlexGen & \multicolumn{1}{c}{} 1.761 & 3.488 & 6.646 & 12.826 & 23.795 & 41.210 \\ \hline
\begin{tabular}[c]{@{}l@{}}{\paperName} (w/o. hiding \\  KV recomputation)\end{tabular} & \multicolumn{1}{c}{} 1.749 & 3.461 & \textbf{6.766} & 12.930 & 23.613 & 43.462 \\ \hline
\begin{tabular}[c]{@{}l@{}}{\paperName} (w. hiding \\ KV recomputation)\end{tabular} & \multicolumn{1}{c}{} \textbf{1.774} & \textbf{3.586} & 6.696 & \textbf{12.986} & \textbf{24.557} & \textbf{43.945}  \\ \hline
\end{tabular}%
}
\caption{OPT-6.7B model with prompt and generation lengths of 256 and 64, respectively. Each MHA block ($W_Q$, $W_K$, $W_V$, and $W_O$) requires 128 MB of memory.}
\label{tab:ablation-hiding}
\end{table}

\noindent \textbf{Runtime breakdown.}
Figure \ref{fig:ablation-runtime} presents the runtime breakdown of an MHA block in {\paperName} and FlexGen during the decoding stage. {\paperName} achieves a substantial reduction in KV cache transfer time, decreasing it from 58\% to 38\%, with activation transfer contributing only 8\% of the total runtime. By recomputing the partial KV cache from the transferred activations, GPU computation time increases from 2.3\% to 13.3\%. This demonstrates that {\paperName} effectively overlaps GPU computation with CPU-GPU communication, substantially reducing the data transfer volume from CPU to GPU and alleviating the PCIe bottleneck that limits LLM inference performance.

\begin{figure}[h]
\centering
    \begin{tikzpicture}
    \node[anchor=east] at (-0.1,1.0) {\small FlexGen};
    \node[anchor=east] at (-0.1,0.5) {\small {\paperName}};

    \begin{axis}[
        xbar stacked,
        bar width=10pt,
        width=0.95\columnwidth, 
        height=2.5cm,
        tick label style={font=\scriptsize},
        ytick=data,
        xtick={0,20,40,60,80,100},
        minor x tick num=4,
        xlabel={Per-unit percentage},
        xlabel style={yshift=5pt},
        xmin=0, xmax=100,
        enlarge y limits=0.5,
        axis x line*=bottom,
        axis y line=none,
        every axis plot/.append style={fill opacity=0.8, draw=none},
        legend entries={load-weight, load-cache, load-activation, compute, store-cache, store-activation},
        legend style={at={(0.5, 1.5)}, anchor=south, font=\scriptsize, row sep=0cm, column sep=0.1cm, legend columns=3, legend cell align=left},
        legend image code/.code={\draw[yshift=-0.06cm] rectangle (0.2cm,0.2cm);},
        ]

        \addplot[fill=gray] coordinates {(31,1)}; 
        \addplot[fill=darkteal] coordinates {(38,1)}; 
        \addplot[fill=yellow!80!black] coordinates {(8.12,1)}; 
        \addplot[fill=red!50!black] coordinates {(13.32,1)}; 
        \addplot[fill=green!50!black] coordinates {(8.98,1)}; 
        \addplot[fill=black] coordinates {(0.20,1)}; 
    \end{axis}

    \begin{axis}[
        xbar stacked,
        bar width=10pt,
        width=0.95\columnwidth, 
        height=3.5cm,
        ytick=data,
        tick label style={font=\scriptsize},
        xmin=0, xmax=100,
        enlarge y limits=0.5,
        axis x line*=bottom,
        axis y line=none,
        every axis plot/.append style={fill opacity=0.8, draw=none},
        ]
        \addplot[fill=gray] coordinates {(30,1)};
        \addplot[fill=darkteal] coordinates {(58,1)};
        \addplot[fill=yellow!80!black] coordinates {(0.254,1)};
        \addplot[fill=red!50!black] coordinates {(2.28,1)};
        \addplot[fill=green!50!black] coordinates {(8.73,1)};
        \addplot[fill=black] coordinates {(0.19,1)};
    \end{axis}

    \end{tikzpicture}

\caption{Runtime breakdown of {\paperName} and FlexGen.}
\label{fig:ablation-runtime}
\end{figure}

\section{Related Works}

To address the memory demands of LLMs in resource-constrained settings, offloading techniques aim to minimize the latency of data transfer between CPUs and GPUs. FlexGen \cite{sheng2023flexgen} proposes to offload weights, activations, and KV cache to CPU memory or external storage and maximizes throughput for larger batch sizes by formulating the optimization as a graph traversal problem. HeteGen \cite{zhao2024hetegen} uses the CPU for partial computation on offloaded weights while transferring the remaining workload to the GPU. TwinPilots \cite{yu2024twinpilots} further optimizes workload balancing between the CPU and GPU at the operator level. 
FastDecode \cite{he2024fastdecode} reduces KV cache data movement by offloading the KV cache and attention computation entirely to the CPU. 
\citeauthor{park2024improving} and Neo \cite{jiang2024neo} overlap GPU linear projection computations with CPU-based attention computations across multiple batches to improve resource utilization.

ALISA \cite{zhao2024alisa} compresses the KV cache based on sparsity and offloads KV cache exceeding GPU memory capacity. When loading the KV cache to the GPU, ALISA recomputes a portion of the KV cache first and then transfers the remainder, where we propose overlapping the recomputation and transfer by adaptively determining the optimal split point. Furthermore, ALISA addresses only the row-by-row schedule, while {\paperName} extends to the column-by-column schedule.
{\paperName} is orthogonal to CPU-assisted and KV cache compression approaches, making it compatible for integration with these techniques to further improve overall system performance. As shown in the additional experiments provided in Appendix \ref{app:comparing-cpu}, we demonstrate that the CPU can become a bottleneck in certain distributed system configurations. In contrast, {\paperName} optimizes GPU utilization and data transfer efficiency without relying on additional CPU resources or approximations of the KV cache.

\section{Conclusion}

In this paper, we introduce {\paperName}, an efficient CPU-GPU I/O-aware LLM inference method designed to accelerate KV cache loading. {\paperName} minimizes the data transfer between the CPU and GPU by leveraging KV cache partial recomputation. By overlapping this recomputation with data transmission, {\paperName} significantly reduces idle GPU time and enhances overall inference performance. 
Future work could extend our method to tolerate KV cache loading from remote network storage or scale to large multi-GPU infrastructure, further enhancing its applicability and performance in diverse deployment scenarios.

\section{Limitations}
Our study represents an important step towards optimizing the efficiency of LLM inference by leveraging KV cache partial recomputation. However, {\paperName} has certain limitations that suggest avenues for future research.
First, our methodology is currently limited to single-GPU and data-parallel multi-GPU inference. It does not yet extend to advanced distributed systems, such as model or tensor parallelism. Expanding this approach to these paradigms could enable support for larger model sizes.
Second, while we address PCIe bandwidth bottlenecks in CPU-GPU communication, we do not consider scenarios where the KV cache is loaded from disk or network storage. Nevertheless, {\paperName} could potentially be adapted to accelerate the prefilling stage in such setups.
Third, the current implementation performs system profiling only at the start of inference, assuming static hardware conditions throughout the process. Incorporating dynamic profiling and runtime adaptive optimization could enhance the robustness and efficiency of the approach, particularly in heterogeneous or multi-tenant environments.

\section{Acknowledgment}
We sincerely thank all the reviewers for their time and constructive comments. This material is based upon work supported by  NSF award number  2224319, REAL@USC-Meta center, and VMware gift. The views, opinions, and/or findings expressed are those of the author(s) and should not be interpreted as representing the official views or policies of the  U.S. Government.


\bibliography{custom}

\clearpage
\appendix
\section{Appendix}

\subsection{Scheduling Methods}
\label{app:schedule}

\begin{figure}[h]
\centering
\begin{subfigure}[t]{0.9\columnwidth}
\centering
    \includegraphics[width=\columnwidth]{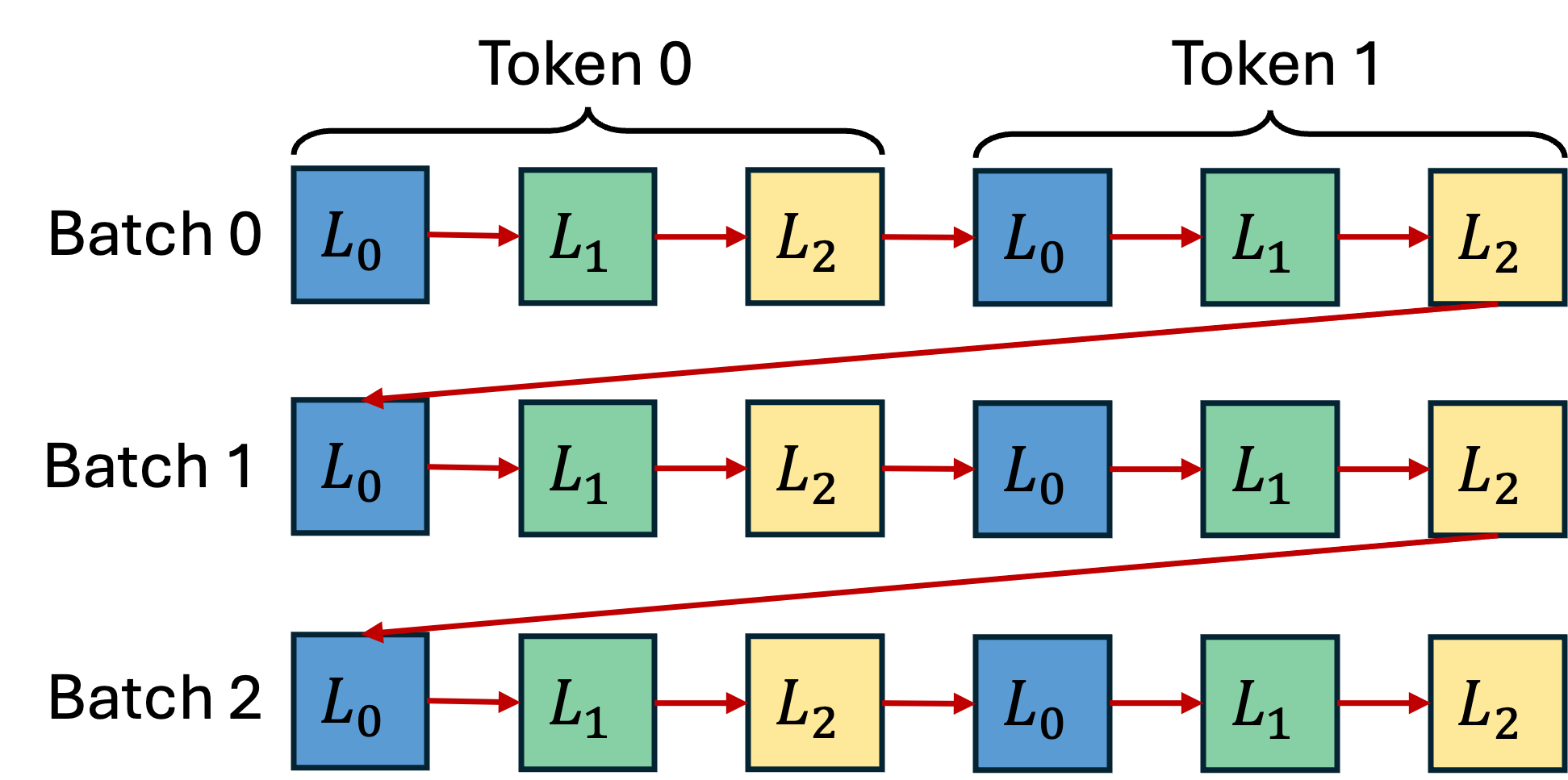}
\caption{row-wise scheduling.}
\end{subfigure}

\begin{subfigure}[t]{0.9\columnwidth}
\centering
    \includegraphics[width=\columnwidth]{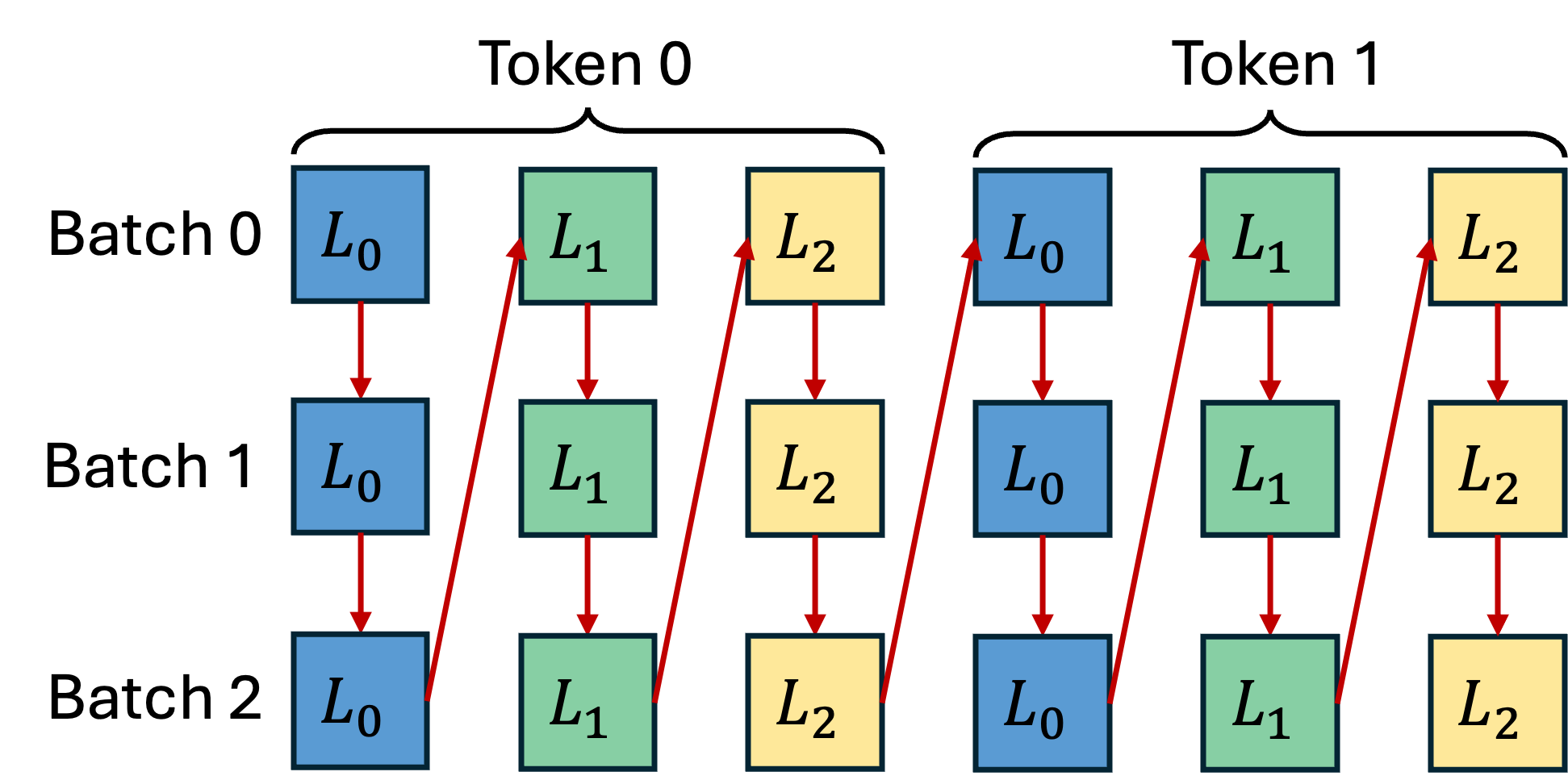}
\caption{column-wise scheduling.}
\end{subfigure}

\caption{Two different scheduling methods, with arrows indicating the scheduling order.}
\label{app:fig:schedule-methods}
\end{figure}

Figures \ref{app:fig:schedule-methods} illustrates two decoding schedules for generating 2 tokens from a model with three layers ($L_0$, $L_1$, and $L_2$) during the decoding stage. In Figure \ref{app:fig:schedule-methods}(a), the row-by-row schedule processes each batch across all layers before moving to the next batch. 
In contrast, Figure \ref{app:fig:schedule-methods}(b) shows the column-by-column schedule, where each layer is reused to process a group of batches before moving to the next layer.

\subsection{KV Cache Partial Recomputation with Overlapping}
\label{app:overlapping}

Built on FlexGen \cite{sheng2023flexgen}'s computation and communication overlapping technique, we adapt it to support KV cache partial recomputation. Algorithm \ref{alg:block_schedule_overlap} enables simultaneous execution of tasks within the innermost loop, including loading weights for the next layer, loading activations for KV cache recomputation, recomputing the partial KV cache, loading the rest of the KV cache and activations for the next batch, storing the KV cache and activations for the previous batch, and performing computation for the current batch. Although the algorithm is designed for column-by-column scheduling, the row-by-row schedule with a single batch is a special case of it.

\begin{algorithm}[t!]
\caption{KV Cache Partial Recomputation with Overlapping}
\label{alg:block_schedule_overlap}
\begin{algorithmic} 
\FOR{$i = 1$ \textbf{to} \texttt{generation\_length}}
    \FOR{$j = 1$ \textbf{to} \texttt{num\_layers}}
        \FOR{$k = 1$ \textbf{to} \texttt{num\_GPU\_batches}}
            \STATE // Load the weight of the next layer
            \STATE \texttt{load\_weight}$(i, j+1, k)$
            
            \STATE // Load the activation for KV cache recomputation of the next batch
            \STATE \texttt{load\_activation\_recompute}$(i, j, k+1)$
            
            \STATE // Load the KV cache and activation of the next batch
            \STATE \texttt{load\_cache}$(i, j, k+1)$
            \STATE \texttt{load\_activation}$(i, j, k+1)$
            
            \STATE // Compute this batch
            \STATE \texttt{compute}$(i, j, k)$
            
            \STATE // Store the KV cache and activation of the previous batch
            \STATE \texttt{store\_activation}$(i, j, k-1)$
            \STATE \texttt{store\_cache}$(i, j, k-1)$
            
            \STATE // Synchronize all devices
            \STATE \texttt{synchronize}$()$
        \ENDFOR
    \ENDFOR
\ENDFOR
\end{algorithmic}
\end{algorithm}

\subsection{Detailed Experimental Results}
\label{app:detailed-exp}
Table \ref{app:tab:opt-6.7b} and \ref{app:tab:opt-13b} present detailed experimental results for latency-oriented workloads using OPT-6.7B and OPT-13B. The results show the performance differences between {\paperName} and the baseline (Hugging Face Transformer with KV cache offloading) in terms of GPU peak memory, decode latency, and throughput across various configurations. Notably, {\paperName} consistently achieves lower latency while maintaining comparable memory usage.

\begin{table*}[h]
\centering
\resizebox{\textwidth}{!}{%
\begin{tabular}{cccccccc}
\hline
\multicolumn{1}{l}{Method} & Batch size & \begin{tabular}[c]{@{}c@{}}Prompt\\ length\end{tabular} & \begin{tabular}[c]{@{}c@{}}Generation \\ length\end{tabular} & \begin{tabular}[c]{@{}c@{}}Cache size \\ (GB)\end{tabular} & \begin{tabular}[c]{@{}c@{}}GPU peak mem \\ (GB)\end{tabular} & \begin{tabular}[c]{@{}c@{}}Decode latency\\ (sec)\end{tabular} & \begin{tabular}[c]{@{}c@{}}Decode throughput\\ (tokens/s)\end{tabular} \\ \hline
\multirow{6}{*}{Accel.} & 64 & 128 & 32 & 5.0 & 14.427 & 8.905 & 222.788 \\ \cline{2-8} 
 & 64 & 128 & 128 & 8.0 & 14.708 & 71.327 & 113.954 \\ \cline{2-8} 
 & 64 & 256 & 32 & 9.0 & 16.337 & 26.825 & 73.961 \\ \cline{2-8} 
 & 64 & 256 & 128 & 12.0 & 16.618 & 88.354 & 91.993 \\ \cline{2-8} 
 & 64 & 512 & 32 & 17.0 & 20.154 & 24.390 & 81.344 \\ \cline{2-8} 
 & 64 & 512 & 128 & 20.0 & 20.576 & 110.277 & 73.705 \\ \hline
\multirow{6}{*}{{\paperName}} & 64 & 128 & 32 & 5.0 & 14.364 & 6.651 & 298.284 \\ \cline{2-8} 
 & 64 & 128 & 128 & 8.0 & 14.645 & 45.766 & 177.598 \\ \cline{2-8} 
 & 64 & 256 & 32 & 9.0 & 16.212 & 19.138 & 103.666 \\ \cline{2-8} 
 & 64 & 256 & 128 & 12.0 & 16.493 & 61.597 & 131.955 \\ \cline{2-8} 
 & 64 & 512 & 32 & 17.0 & 19.904 & 20.349 & 97.501 \\ \cline{2-8} 
 & 64 & 512 & 128 & 20.0 & 20.951 & 93.932 & 86.531 \\ \hline
\end{tabular}
}
\caption{Detailed experimental results for OPT-6.7B corresponding to Figure \ref{fig:latency-exp}.}
\label{app:tab:opt-6.7b}
\end{table*}

\begin{table*}[h]
\centering
\resizebox{\textwidth}{!}{%
\begin{tabular}{cccccccc}
\hline
\multicolumn{1}{l}{Method} & Batch size & \begin{tabular}[c]{@{}c@{}}Prompt\\ length\end{tabular} & \begin{tabular}[c]{@{}c@{}}Generation \\ length\end{tabular} & \begin{tabular}[c]{@{}c@{}}Cache size \\ (GB)\end{tabular} & \begin{tabular}[c]{@{}c@{}}GPU peak mem \\ (GB)\end{tabular} & \begin{tabular}[c]{@{}c@{}}Decode latency\\ (sec)\end{tabular} & \begin{tabular}[c]{@{}c@{}}Decode throughput\\ (tokens/s)\end{tabular} \\ \hline
\multirow{6}{*}{Accel.} & 64 & 128 & 32 & 7.812 & 26.083 & 11.409 & 173.891  \\ \cline{2-8} 
 & 64 & 128 & 128 & 12.500 & 26.434 & 73.896 & 109.993 \\ \cline{2-8} 
 & 64 & 256 & 32 & 14.062 & 28.087 & 19.381 & 102.368 \\ \cline{2-8} 
 & 64 & 256 & 128 & 18.750 & 28.439 & 104.115 & 78.068 \\ \cline{2-8} 
 & 64 & 512 & 32 & 26.562 & 32.851 & 35.066 & 56.579 \\ \cline{2-8} 
 & 64 & 512 & 128 & 31.250 & 34.146 & 168.155 & 48.336 \\ \hline
\multirow{6}{*}{{\paperName}} & 64 & 128 & 32 & 7.812 & 26.005 & 9.148 & 216.867 \\ \cline{2-8} 
 & 64 & 128 & 128 & 12.500 & 26.356 & 66.119 & 122.929 \\ \cline{2-8} 
 & 64 & 256 & 32 & 14.062 & 27.931 & 16.654 & 119.127 \\ \cline{2-8} 
 & 64 & 256 & 128 & 18.750 & 28.337 & 88.492 & 91.850 \\ \cline{2-8} 
 & 64 & 512 & 32 & 26.562 & 33.203 & 29.215 & 67.911 \\ \cline{2-8} 
 & 64 & 512 & 128 & 31.250 & 34.615 & 138.377 & 58.738 \\ \hline
\end{tabular}
}
\caption{Detailed experimental results for OPT-13B corresponding to Figure \ref{fig:latency-exp}.}
\label{app:tab:opt-13b}
\end{table*}

\subsection{Optimal KV Cache Split Points}
\label{app:optimal-split-points}

Figure \ref{app:fig:split-points} presents the optimal KV cache split points $l$, obtained by solving the linear programming problem defined in Eq.~\eqref{eq:lp}, for the first setting of the latency-oriented workload experiments in Section \ref{sec:exp} (prompt length of 128 and generation length of 32). Based on system profiling statistics and KV cache size, the optimal split point $l$ is 182 when the generation length is 1, and $l$ increases to 128 when the generation length is 32.

\begin{figure}[h]
\centering
\begin{tikzpicture}
    \begin{axis}[
        axis line style={line width=1pt},
        width=0.95\columnwidth,
        height=4.5cm,
        xlabel={Number of tokens generated},
        ylabel={Recomputation length $l$},
        ylabel style={yshift=-10pt},
        tick label style={font=\small},
        grid=both,
        xtick={1,4,8,12,16,20,24,28,32}, 
        ytick={180,185,190,195,200,205}, 
    ]
    \addplot[color=darkteal, thick] coordinates {
        (1, 182) (2, 183) (3, 184) (4, 184) (5, 185) (6, 186) 
        (7, 186) (8, 187) (9, 188) (10, 188) (11, 189) (12, 190) 
        (13, 191) (14, 191) (15, 192) (16, 193) (17, 193) 
        (18, 194) (19, 195) (20, 196) (21, 196) (22, 197) 
        (23, 198) (24, 198) (25, 199) (26, 200) (27, 201) 
        (28, 201) (29, 202) (30, 203) (31, 203) (32, 204)
    };
    \end{axis}
\end{tikzpicture}
\caption{Optimal KV cache split points $l$ over the generation process.}
\label{app:fig:split-points}
\end{figure}

\subsection{System Performance with a Low-end GPU}
\label{app:low-end-gpu-experiments}

To further demonstrate the adaptability of {\paperName}, we evaluate it on a low-end system with an AMD EPYC 32-Core CPU and an NVIDIA Quadro RTX 5000 GPU (16 GB HBM, 89.2 TFLOPS FP16 peak performance) connected via PCIe 4.0 x8 (16 GB/s bandwidth). GPU TFLOPS, GPU memory, and PCIe bandwidth are lower in this system setting than those in the default system we used earlier. Despite the reduced GPU speed and bandwidth, {\paperName} achieves up to 15\% higher throughput than FlexGen for OPT-6.7B in the same throughput-oriented workload, as shown in Table \ref{tab:throughput_exp}.

\begin{table}[h]
\centering
\resizebox{\columnwidth}{!}{%
\begin{tabular}{ccccccc}
\hline
\textbf{Seq len} & \textbf{256/32} & \textbf{256/128} & \textbf{512/32} & \textbf{512/128} & \textbf{1024/32} & \textbf{1024/128} \\ \hline
 FlexGen & 50.057 & 46.779 & 29.614 & 28.650 & 15.778 & 16.194 \\ \hline
 {\paperName}    & 53.976 & 49.860 & 33.666 & 32.277 & 18.285 & 18.108 \\ \hline
\end{tabular}
}
\caption{Throughput (tokens/s) comparison on a low-end GPU system.}
\label{tab:throughput_exp}
\end{table}

\subsection{Additional Experimental Results on LLaMa Models}

In addition to the OPT models discussed in Section~\ref{sec:exp}, we conduct further experiments on the more recent LLaMa2-7B and LLaMa2-13B models. Using the same experimental setup, we measure decoding throughput while processing a single batch of size 64 across varying prompt and generation lengths. As shown in Figure~\ref{fig:llama-exp}, {\paperName} consistently achieves higher throughput than the baselines (DeepSpeed Inference and Hugging Face Accelerate) on both LLaMa2 models.

\begin{figure}[h]
\centering
\begin{subfigure}{\columnwidth}
    \centering
    \begin{tikzpicture}
        \begin{axis}[
            axis line style={line width=1pt},
            title={Llama2-7B},
            title style={yshift=-5pt},
            width=\columnwidth,
            height=5cm,
            ybar=0pt,
            bar width=.2cm,
            enlarge x limits=0.15,
            tick label style={font=\scriptsize},
            symbolic x coords={A, B, C, D, E, F},
            xticklabel style={align=center},
            xticklabels={{256 \\ 32}, {256 \\ 128}, {512 \\ 32}, {512 \\ 128}, {1024 \\ 32}, {1024 \\ 128}},
            xtick=data,
            xtick style={draw=none},
            xticklabel style={yshift=5pt, align=center},
            xlabel={Sequence length},
            ylabel={Throughput (tokens/s)},
            ylabel style={yshift=-10pt},
            ymin=0,
            ytick={0, 50, 100, 150, 200, 250},
            grid=major,
            grid style=solid,
            major x grid style={draw=none},
            legend entries={DS, Accel., {\paperName}},
            legend style={at={(1,1)}, anchor=north east, font=\scriptsize, legend cell align=center, legend columns=3},
            legend image code/.code={\draw[yshift=-0.06cm] rectangle (0.2cm,0.2cm);},
        ]

        \addplot[fill=black] coordinates {(A,162.748) (B,124.532) (C,80.215) (D,69.843) (E,40.927) (F,36.482)};

        \addplot[fill=gray] coordinates {(A,166.545) (B,132.054) (C,94.122) (D,80.048) (E,48.599) (F,44.453)};

        \addplot[fill=darkteal] coordinates {(A,217.822) (B,195.822) (C,126.797) (D,115.597) (E,71.164) (F,66.856)};

        \end{axis}
    \end{tikzpicture}
\end{subfigure}

\begin{subfigure}{\columnwidth}
    \centering
    \begin{tikzpicture}
        \begin{axis}[
            axis line style={line width=1pt},
            title={Llama2-13B},
            title style={yshift=-5pt},
            width=\columnwidth,
            height=5cm,
            ybar=0pt,
            bar width=.2cm,
            enlarge x limits=0.15,
            tick label style={font=\scriptsize},
            symbolic x coords={A, B, C, D, E, F},
            xticklabel style={align=center},
            xticklabels={{256 \\ 32}, {256 \\ 128}, {512 \\ 32}, {512 \\ 128}, {1024 \\ 32}, {1024 \\ 128}},
            xtick=data,
            xtick style={draw=none},
            xticklabel style={yshift=5pt, align=center},
            xlabel={Sequence length},
            ylabel={Throughput (tokens/s)},
            ylabel style={yshift=-10pt},
            ymin=0,
            ytick={0, 50, 100, 150, 200},
            grid=major,
            grid style=solid,
            major x grid style={draw=none},
            legend entries={DS, Accel., {\paperName}},
            legend style={at={(1,1)}, anchor=north east, font=\scriptsize, legend cell align=center, legend columns=3},
            legend image code/.code={\draw[yshift=-0.06cm] rectangle (0.2cm,0.2cm);},
        ]

        \addplot[fill=black] coordinates {(A,104.382) (B,75.129) (C,46.735) (D,42.647) (E,22.914) (F,19.526)};

        \addplot[fill=gray] coordinates {(A,111.248) (B,84.130) (C,59.900) (D,50.966) (E,30.911) (F,28.360)};

        \addplot[fill=darkteal] coordinates {(A,163.078) (B,113.390) (C,80.158) (D,70.169) (E,43.429) (F,41.284)};

        \end{axis}
    \end{tikzpicture}
\end{subfigure}

\caption{Decoding throughput for a single batch of size 64 across different sequence lengths.}
\label{fig:llama-exp}
\end{figure}

\subsection{Comparing with CPU-assisted Approaches in Distributed System Setup}
\label{app:comparing-cpu}

In this experiment, we compare the performance of the CPU-assisted offloading approach, FastDecode \cite{he2024fastdecode}, with {\paperName} on a GPU node equipped with 8 NVIDIA A100 GPUs and a single CPU, which is the same AMD EPYC processor (64 cores with PCIe 4.0 128 lanes), as described in Section \ref{sec:exp}.

We run multiple concurrent processes of FastDecode and {\paperName} on the available GPUs, with each GPU dedicated to a single process. This setup simulates scenarios where either multiple users share a single computing node or a single user performs data-parallel inference. FastDecode relies on the CPU for attention computations, resulting in a performance drop as the CPU becomes a bottleneck when managing multiple concurrent inference processes. In contrast, {\paperName} eliminates CPU dependency entirely and instead optimizes data transfer over the PCIe bus. 

Figure \ref{app:fig:cpu} demonstrates that while FastDecode suffers a significant decline in throughput as the number of processes increases, {\paperName} exhibits better scalability, maintaining stable performance in systems with a single CPU and multiple GPUs.

\begin{figure}[h]
\centering
\begin{tikzpicture}
    \begin{axis}[
        axis line style={line width=1pt},
        width=0.95\columnwidth,
        height=4.5cm,
        xlabel={Number of concurrent GPU processes},
        ylabel={Throughput (tokens/s)},
        ylabel style={yshift=-10pt},
        xtick={1,2,3,4},
        xticklabels={1,2,4,8},
        xtick style={draw=none},
        grid=both,
        major grid style={dashed},
        minor grid style={dotted},
        tick label style={font=\small},
        legend entries={{\paperName}, FastDecode},
        legend style={at={(0,0)}, anchor=south west, font=\small, row sep=0cm, column sep=0cm, legend cell align=left},
    ]
    \addplot[color=darkteal, mark=square*, thick] coordinates {
        (1, 41.343)
        (2, 41.026)
        (3, 36.626)
        (4, 36.069)
    };    
    \addplot[color=gray, mark=*, thick] coordinates {
        (1, 42.579)
        (2, 40.263)
        (3, 28.354)
        (4, 19.903)
    };
    \end{axis}
\end{tikzpicture}
\caption{Throughput comparison between {\paperName} and FastDecode with different GPU workload.}
\label{app:fig:cpu}
\end{figure}

\subsection{Extended Related Works}
\textbf{GPU-efficient LLM inference.} 
Maximizing GPU utilization is crucial for serving LLMs efficiently to achieve low latency and high throughput.
Orca \cite{yu2022orca} employs iteration-level scheduling to handle batches with varying output lengths, returning completed sequences immediately to serve new ones. 
PagedAttention \cite{kwon2023vllm} observes that the KV cache grows and shrinks dynamically as tokens are generated, though the sequence lifetime and length are not predetermined. It addresses this by managing the KV cache as non-contiguous memory blocks. FlashAttention \cite{dao2022flashattention} combines attention operations into a single kernel and tiles QKV matrices into smaller blocks to optimize GPU SRAM usage and reduce HBM access overhead, while our work mainly focuses on optimizing PCIe bandwidth. DeepSpeed-Inference \cite{aminabadi2022deepspeed} enhances multi-GPU inference for both dense and sparse Transformer models by combining GPU memory and employing a hybrid inference technique with CPU and NVMe memory. Flash-Decoding \cite{flashdecoding2023} accelerates long-context inference by splitting keys and values into smaller chunks, enabling parallel attention computations and combining results for the final output. HCache \cite{gao2025fast} focuses on restoring contextual states across user requests for reuse in online inference, balancing latency, capacity, and persistence for robust LLM serving.

\noindent \textbf{KV cache optimization.} 
Efficient KV cache management enhances inference performance through compression or eviction strategies. KIVI \cite{liu2024kivi} introduces a tuning-free 2-bit quantization method to compress key cache per channel and value cache per token. Similarly, KVQuant \cite{hooper2024kvquant} applies 3-bit compression by combining per-channel quantization with pre-rotary positional embedding quantization for LLaMA. For eviction, H2O \cite{zhang2023h2o} formulates KV cache eviction as a dynamic submodular problem, prioritizing critical and recent tokens to improve throughput. StreamingLLM \cite{xiao2023efficient} uses window attention with a fixed-size sliding window to retain the most recent KV caches, maintaining constant memory usage and decoding speed once the cache reaches capacity. InfiniGen \cite{lee2024infinigen} stores low-rank key cache in GPU memory, offloads value cache to the CPU, and selectively retrieves important values based on approximate attention scores.

\end{document}